%% file: InteractiveModelAnalysis_paper.tex
\documentclass[review]{elsarticle}

\usepackage{hyperref}
\usepackage{lineno}
\usepackage{paralist}
\usepackage{color}
\modulolinenumbers[5]

\journal{Journal of \LaTeX\ Templates}

\newcommand{\doc}[1]{\textcolor{black}{#1}}

\begin{document}

\begin{frontmatter}

\title{Towards Better Analysis of Machine Learning Models: A Visual Analytics Perspective}
\tnotetext[mytitlenote]{Fully documented templates are available in the elsarticle package on \href{http://www.ctan.org/tex-archive/macros/latex/contrib/elsarticle}{CTAN}.}

%% Group authors per affiliation:
\author{Shixia Liu, Xiting Wang, Mengchen Liu, Jun Zhu}
\address{Tsinghua University, Beijing, China}
%\fntext[myfootnote]{Since 1880.}

%% or include affiliations in footnotes:
%\author[mymainaddress,mysecondaryaddress]{Elsevier Inc}
%\ead[url]{www.elsevier.com}
%
%\author[mysecondaryaddress]{Global Customer Service\corref{mycorrespondingauthor}}
%\cortext[mycorrespondingauthor]{Corresponding author}
%\ead{support@elsevier.com}
%
%\address[mymainaddress]{1600 John F Kennedy Boulevard, Philadelphia}
%\address[mysecondaryaddress]{360 Park Avenue South, New York}

\begin{abstract}
Interactive model analysis, the process of understanding, diagnosing, and refining a machine learning model with the help of interactive visualization, is very important for users to efficiently solve real-world artificial intelligence and data mining problems.
%Dramatic success in big data analytics has led to a torrent of interactive model analysis tasks.
Dramatic \doc{advances} in big data analytics has led to a wide variety of interactive model analysis tasks.
In this paper, we present a comprehensive analysis and interpretation of this rapidly developing area.
Specifically, we classify the relevant work into three categories: understanding, diagnosis, and refinement.
%Each category is exemplified by the recent influential work.
Each category is exemplified \doc{by recent} influential work.
Possible future research opportunities are also explored and discussed.
\end{abstract}


\end{frontmatter}

%\linenumbers

\input{intro.tex}
\input{scope.tex}
\input{status.tex}
\input{challenges.tex}

\section*{References}

\bibliography{mybibfile}

\end{document}

%% file: intro.tex
\section{Introduction}

%The widespread success of machine learning has led to numerous real-world applications.
%Machine learning has demonstrated being highly successful in many real-world applications
%Machine learning has been successfully applied to a wide variety of fields ranging from information retrieval, data mining, speech recognition, to computer graphics, visualization, and human-computer interaction.
Machine learning has been successfully applied to a wide variety of fields ranging from information retrieval, data mining, \doc{and} speech recognition, to computer graphics, visualization, and human-computer interaction.
%However, most users often treat a machine learning model as a black box because of its incomprehensible functions and unclear working mechanism~\cite{Fekete2013_Computer_Visual,Liu2016_TVCG_Towards,Muhlbacher2014_TVCG_Opening}.
However, most users often treat a machine learning model as a black box because of its incomprehensible functions and unclear working mechanism~\cite{Fekete2013_Computer_Visual,Liu2016_TVCG_Towards,Muhlbacher2014_TVCG_Opening}.
%Without a clear understanding of how and why the model works, the development of high-performance models typically relies on a time-consuming trial-and-error procedure.
Without a clear understanding of how and why \doc{a} model works, the development of high-performance models typically relies on a time-consuming trial-and-error \doc{process}.
%As a result, academic researchers and industrial practitioners are facing the challenges that demand more transparent and explainable systems for better understanding and analyzing machine learning models, especially their working mechanisms.
As a result, academic researchers and industrial practitioners are \doc{facing challenges} that demand more transparent and explainable systems for better understanding and analyzing machine learning models, especially their inner working mechanisms.

%To tackle the aforementioned challenges, there are some initial efforts on interactive model analysis, which demonstrate that interactive visualization plays a critical role in understanding and analyzing a variety of machine learning models.
To tackle the aforementioned challenges, there are some initial efforts on interactive model analysis.
\doc{These efforts have shown} that interactive visualization plays a critical role in understanding and analyzing a variety of machine learning models.
%Recently, DARPA I2O has released the Explainable Artificial Intelligence (XAI)~\cite{DARPA2016} to encourage the research on this topic.
Recently, DARPA \doc{I2O released Explainable} Artificial Intelligence (XAI)~\cite{DARPA2016} to \doc{encourage research} on this topic.
%The main goal of XAI is to create a suite of machine learning techniques that produce explainable models, which enable users to understand, trust, and manage the emerging generation of Artificial Intelligence (AI) systems.
The main goal of XAI is to create a suite of machine learning techniques that produce explainable \doc{models to enable} users to understand, trust, and manage the emerging generation of Artificial Intelligence (AI) systems.

%Based on the roles of the interactive visualization techniques,
In this paper, we first provide an overview of interactive model analysis.
Then we summarize recent interactive model analysis techniques based on their target tasks (such as understanding how a classifier works)~\cite{Heimerl2012_TVCG_Visual}.
Research opportunities and future directions are discussed for developing new interactive model analysis techniques and systems.

%% file: scope.tex
\section{Scope and Overview}

\begin{figure}
\centering
\includegraphics[width=\textwidth]{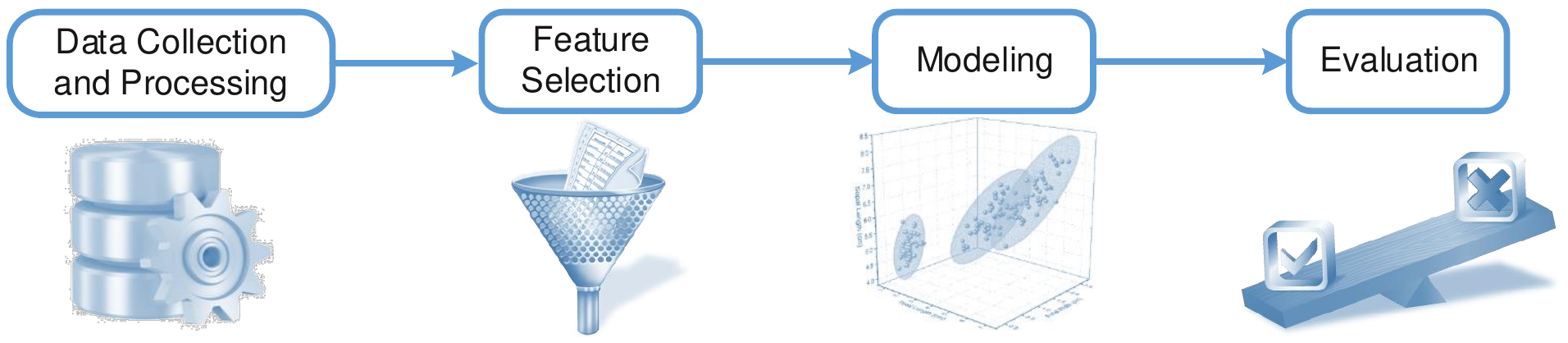}
\caption{A pipeline of machine learning.}
\label{fig.ml_overview}
\end{figure}

%The research and application problems we are focusing on are in the context of machine learning.
\doc{We are focused on} research and application problems \doc{within} the context of machine learning.
Fig.~\ref{fig.ml_overview} illustrates a typical machine learning pipeline, from which we first obtain data.
Then we extract features that are usable as input to a machine learning model.
%Next, the model is trained, tested, and gradually refined based on the evaluation results and the expertise of machine learning experts, which is both time consuming and uncertain in building a reliable model.
Next, the model is trained, tested, and gradually refined based on the evaluation results and \doc{experience} of machine learning experts, \doc{a process that} is both time consuming and uncertain in building a reliable model.
%In addition to an explosion of work on better understanding of learning results~\cite{Cui2011_TVCG_TextFlow,Cui2014_TVCG_How,Dou2013_TVCG_HierarchicalTopics,Dou2016_CGA_Topic,Liu2012_TIST_TIARA,Liu2014_TheVisualComputer_Survey,Liu2016_TVCG_Online,Wang2013_KDD_Mining,Wang2016_TVCG_TopicPanorama}, there has been increasing attentions paid by researchers to leveraging interactive visualization to better understand and iteratively improve a machine learning model.
In addition to an explosion of \doc{research} on better understanding of learning results~\cite{Cui2011_TVCG_TextFlow,Cui2014_TVCG_How,Dou2013_TVCG_HierarchicalTopics,Dou2016_CGA_Topic,Liu2012_TIST_TIARA,Liu2014_TheVisualComputer_Survey,Liu2016_TVCG_Online,Wang2013_KDD_Mining,Wang2016_TVCG_TopicPanorama}, \doc{researchers have paid increasing attention} to leveraging interactive \doc{visualizations} to better understand and iteratively improve a machine learning model.
%The main goal of such research is to reduce human efforts in training a reliable and accurate model.
The main goal of such research is to reduce human \doc{effort when} training a reliable and accurate model.
We refer to the aforementioned iterative and progressive process as interactive model analysis.
%a rapid growth of interest in tightly integrating interactive visualization with machine learning techniques to help  users consume huge amounts of information~\cite{}.
%These efforts can be classified into two groups: understanding of learning results,

\begin{figure}
\centering
\includegraphics[width=\textwidth]{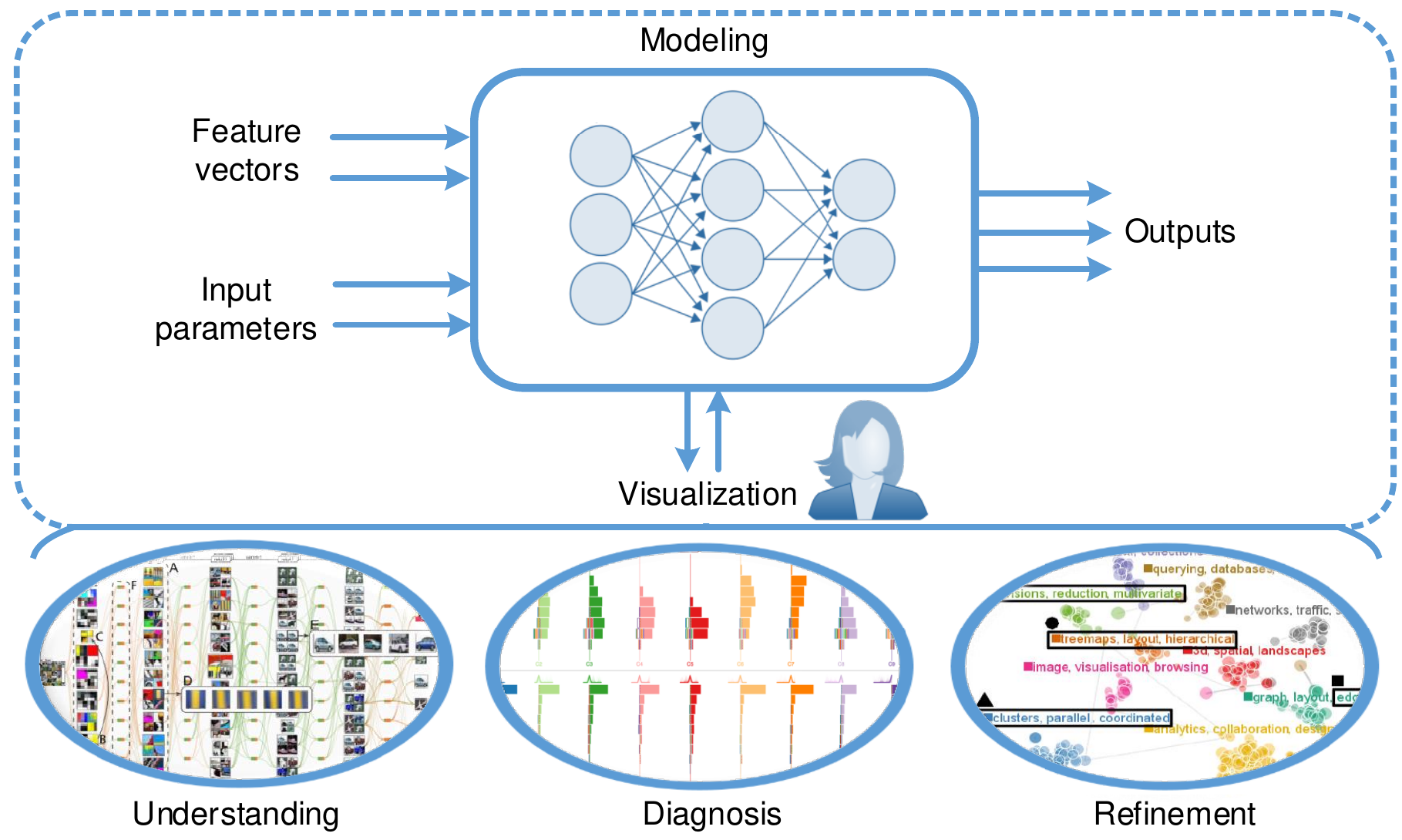}
\caption{An overview of interactive model analysis.}
\label{fig.ima_overview}
\end{figure}

%Fig.~\ref{fig.ima_overview} illustrates the basic idea of interactive model analysis, where machine learning models are seamlessly integrated with state-of-the-art interactive visualization techniques capable of translating models into understandable and useful explanation dialogues for the expert.
Fig.~\ref{fig.ima_overview} illustrates the basic idea of interactive model analysis, where machine learning models are seamlessly integrated with state-of-the-art interactive visualization techniques capable of translating models into understandable and useful \doc{explanations} for \doc{an} expert.
%The strategy is to pursue a variety of visual analytics techniques in order to help experts to understand, diagnose, and refine a machine learning model.
The strategy is to pursue a variety of visual analytics techniques in order to help \doc{experts understand}, diagnose, and refine a machine learning model.
%Accordingly, interactive model analysis aims at creating a suite of visual analytics techniques that
Accordingly, interactive model analysis aims \doc{to create} a suite of visual analytics techniques that
%enables machine learning experts to:

\begin{compactitem}

%\item understand how machine learning models behave the way they do and why they differ from each other (\textbf{understanding});
\item understand \doc{why} machine learning models behave the way they do and why they differ from each other (\textbf{understanding});
\item diagnose a training process that fails to converge or does not achieve an acceptable performance (\textbf{diagnosis});
%\item guide experts to improve the performance and robustness of machine learning models (\textbf{refinement});
\item guide experts to improve the performance and robustness of machine learning models (\textbf{refinement})\doc{.}

\end{compactitem}

%% file: status.tex
\section{Discussion and Analysis of Existing Work}
%Most of the recent efforts on interactive model analysis aim at helping machine learning experts understand how the model works such as the interactions between each component in the model.
Most \doc{recent} efforts \doc{in} interactive model analysis aim \doc{to help} machine learning experts understand how the model works\doc{,} such as the interactions between each component in the model.
%More recently, there are some initial attempts to diagnose a training process that failed to converge or find a potential direction to improve the model.
More recently, there \doc{have been} some initial attempts to diagnose a training process that failed to converge or did not achieve \doc{the} desired performance, or \doc{to} refine the learning model for better performance.

\subsection{Understanding}
Many techniques have been developed to help experts better understand classification models~\cite{Paiva2015_TVCG_Approach,Turner2016_ICMLWorkshop_Model,Tzeng2005_IEEEVisualization_Opening} and regression models~\cite{Zahavy2016_ICML_Graying}.
%Among all the models, neural networks have received the most attention.
Among \doc{all models}, neural networks have received the most attention.
%Neural networks have been widely used and achieved state-of-the-art results in many machine learning tasks such as image classification and video classification~\cite{Lecun2015_Nature_Deep}.
\doc{They} have been widely used and achieved state-of-the-art results in many machine learning \doc{tasks, such} as image classification and video classification~\cite{Lecun2015_Nature_Deep}.
%However, neural networks are mostly used as a black box and users often fail to explain how learning from training samples was done~\cite{Tzeng2005_IEEEVisualization_Opening}.
%To better understand the working mechanism of neural networks, researchers have developed various visualization approaches.
%These approaches can be classified into two categories: point-based and network-based.\looseness=-1
To better understand the working mechanism of neural networks, researchers have developed various visualization \doc{approaches, which} can be classified into two categories: point-based and network-based.

%Point-based methods~\cite{Zahavy2016_ICML_Graying,Rauber2016_TVCG_Visualizing} reveal the relationships between neural network components (e.g. neurons, activations) by displaying components that are highly related adjacent to each other.
%Point-based techniques~\cite{Zahavy2016_ICML_Graying,Rauber2016_TVCG_Visualizing} reveal the relationships between neural network components such as neurons or learned representations, by using scatterplots.
Point-based techniques~\cite{Zahavy2016_ICML_Graying,Rauber2016_TVCG_Visualizing} reveal the relationships between neural network \doc{components, such} as neurons or learned representations, by using scatterplots.
%Here each learned representation is a high-dimensional vector whose entries are the output values of neurons in one hidden layer.
\doc{Each} learned representation is a high-dimensional vector whose entries are the output values of neurons in one hidden layer.
Typically, each component is represented by a point.
Components with similar roles are placed adjacent to each other by using dimension reduction techniques such as Principal Component Analysis (PCA)~\cite{Wold1987_CILS_Principal} and t-SNE~\cite{Maaten2008_JMLR_Visualizing}.
%It has been shown that point-based techniques facilitate the confirmation of hypothesis about neural networks and identification of previously unknown relationships between neural network components~\cite{Rauber2016_TVCG_Visualizing}.
\doc{Point-based} techniques facilitate the confirmation of hypothesis \doc{on} neural networks and \doc{the} identification of previously unknown relationships between neural network components~\cite{Rauber2016_TVCG_Visualizing}.

\begin{figure}
\centering
\includegraphics[width=1\textwidth]{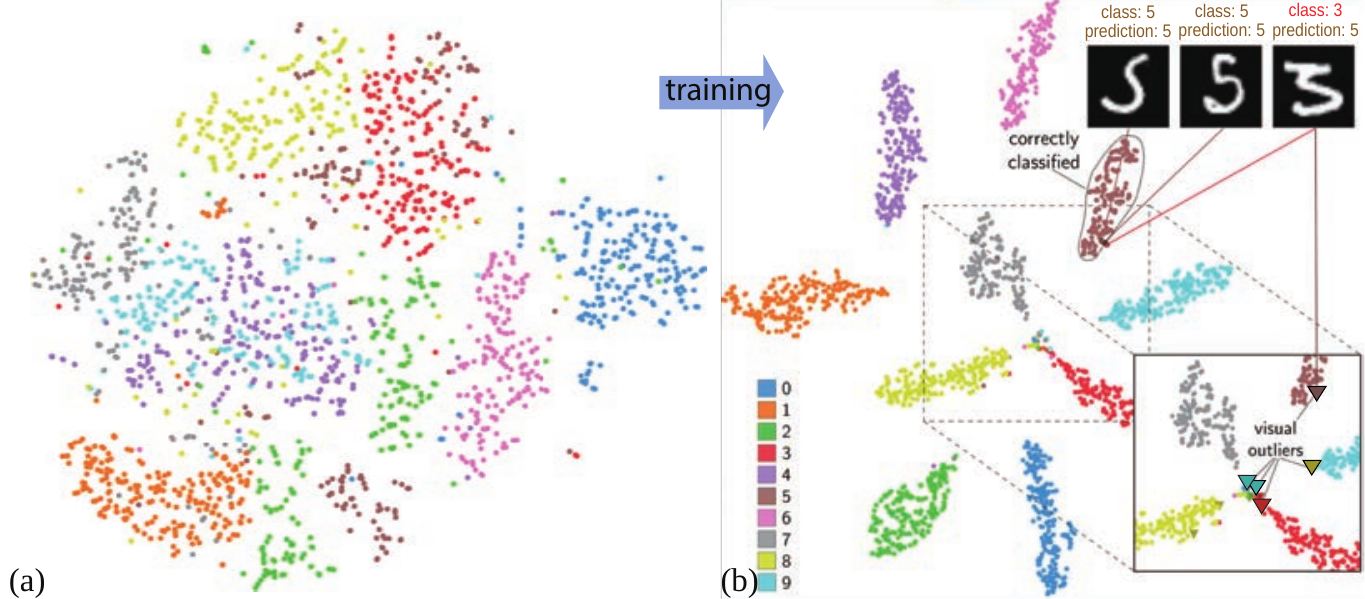}
\caption{Comparison of test sample representations (a) before and (b) after training~\cite{Rauber2016_TVCG_Visualizing}.}
\label{fig:point-based}
\end{figure}

Fig.~\ref{fig:point-based} shows a point-based visualization developed by Rauber et al.~\cite{Rauber2016_TVCG_Visualizing}.
In this figure, each point denotes the learned representation of a test sample.
% in the last hidden layer.
%Here each learned representation is a high-dimensional vector whose entries are the output values of neurons in the last hidden layer.
The color of each point encodes the class label of each test sample.
As shown in the figure, after training, the visual separation between classes is significantly improved.
%This observation provides evidence for the hypothesis that neural networks learn to detect representations useful for class discrimination.
This observation provides evidence for the hypothesis that neural networks learn to detect representations \doc{that are} useful for class discrimination.
%Fig.~\ref{fig:point-based}(b) also helps understand misclassified samples, which are marked by triangle glyphs.
Fig.~\ref{fig:point-based}(b) also helps \doc{with the understanding of} misclassified samples, which are marked by triangle glyphs.
The figure illustrates that many misclassified samples are visual outliers whose neighbors have different classes.
%Also, many outliers correspond to test samples that are even hard for humans to classify.
Also, many outliers correspond to test samples that are \doc{difficult} for \doc{even} humans to classify.
For example, an image of digit 3 is misclassified because it is very similar to some images of digit 5.

\begin{figure}[t]
\centering
\includegraphics[width=\textwidth]{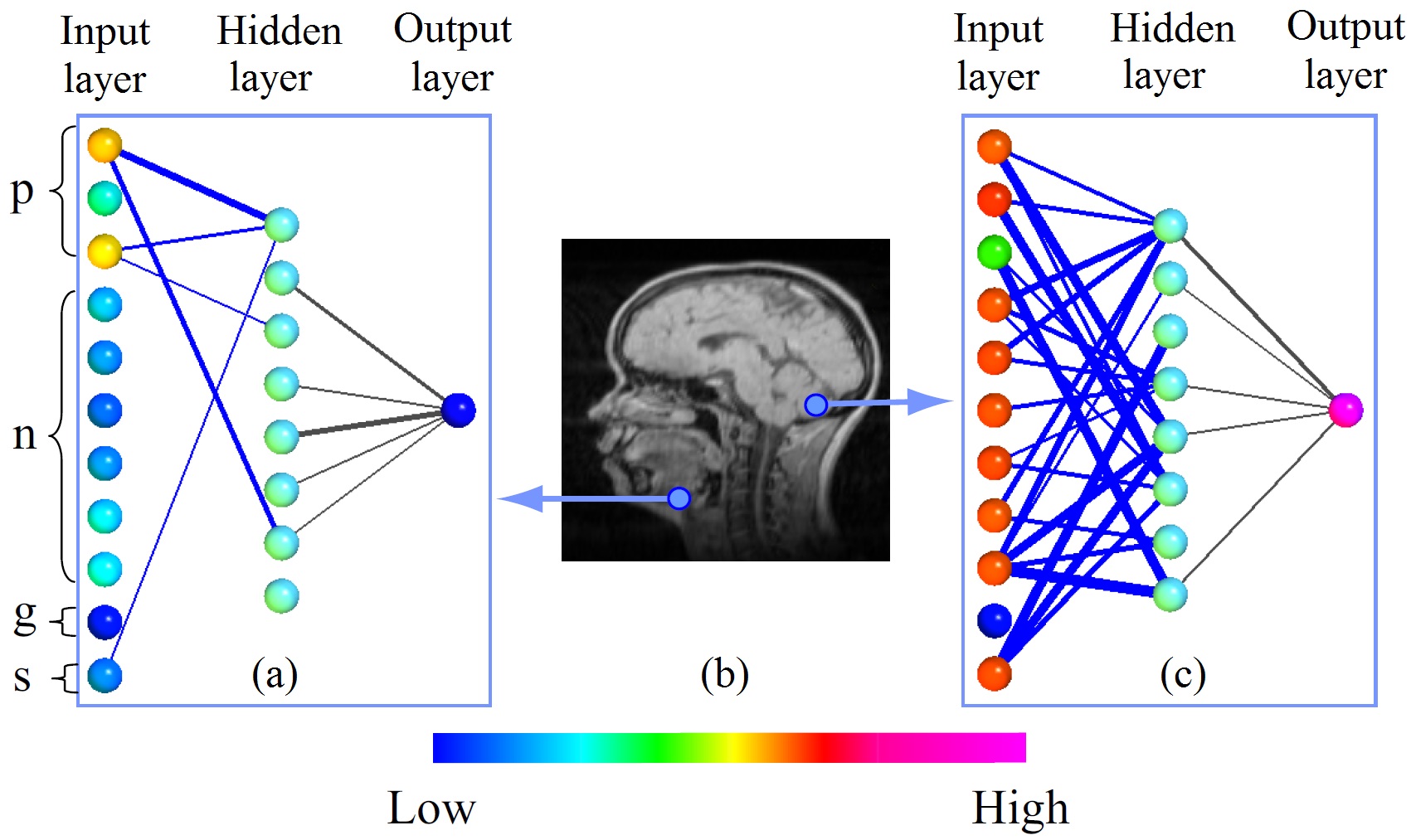}
\caption{Topology of a neural network trained to classify brain and non-brain materials~\cite{Tzeng2005_IEEEVisualization_Opening}.}
\label{fig:network-based}
\end{figure}

Although point-based techniques are useful for presenting the relationships between a large number of neural network components, they cannot reveal the topological information of the networks.
%As a result, they fail to provide a comprehensive understanding of the roles of different neurons in different layers as well as interactions between them.
As a result, they fail to provide a comprehensive understanding of the roles of different neurons in different layers and the interactions between them.
Network-based techniques~\cite{Harley2015_ISVC_Interactive,Streeter2001_SPIE_NVIS,Craven1992_IJAIT_Visualizing} solve this problem by displaying the network topology.
%These techniques usually represent a neural network as a directed acyclic graph (DAG) and encode important information of the network by the size, color, and glyphs of the nodes or edges in the DAG.
These techniques usually represent a neural network as a directed acyclic graph (DAG) and encode important information \doc{from} the network by the size, color, and glyphs of the nodes or edges in the DAG.

Fig.~\ref{fig:network-based} shows the visualization generated by a pioneer network-based technique~\cite{Tzeng2005_IEEEVisualization_Opening}.
%This figure presents a neural network with one hidden layer.
%This neural network is trained to classify whether a voxel within the head belongs to the brain or not.
This figure presents a neural network trained to classify whether a voxel within the head belongs to the brain or not.
Here, each voxel is represented by its scalar value $s$, gradient magnitude $g$, scalar values of its neighbors $n$, and its position $p$. %, which consists of the $x$, $y$, and $z$ coordinates of the voxel.
The width of each edge encodes the importance of the corresponding connection.
The nodes in the input and output layers are colored based on their output values.
%The color of the output node in Fig.~\ref{fig:network-based}(a) and Fig.~\ref{fig:network-based}(c) shows that the neural network is able to correctly classify the voxel on the lower left side and the voxel within the brain.
%From the color of the node in the output layer, we can see that the neural network is able to correctly classify the voxel on the left to non-brain materials (low output value) and the voxel on the right to brain materials (high output value).
\doc{The} color of the node in the output layer \doc{indicates} that the neural network is able to correctly classify the voxel on the left to non-brain materials (low output value) and the voxel on the right to brain materials (high output value).
The network topologies in Fig.~\ref{fig:network-based}(a) and Fig.~\ref{fig:network-based}(c) demonstrate that the voxel on the left is classified to non-brain materials mainly because of its position, while the voxel on the right needs all inputs except for the gradient magnitude $g$ to be correctly classified to brain materials.

\begin{figure}[t]
\centering
\includegraphics[width=\textwidth]{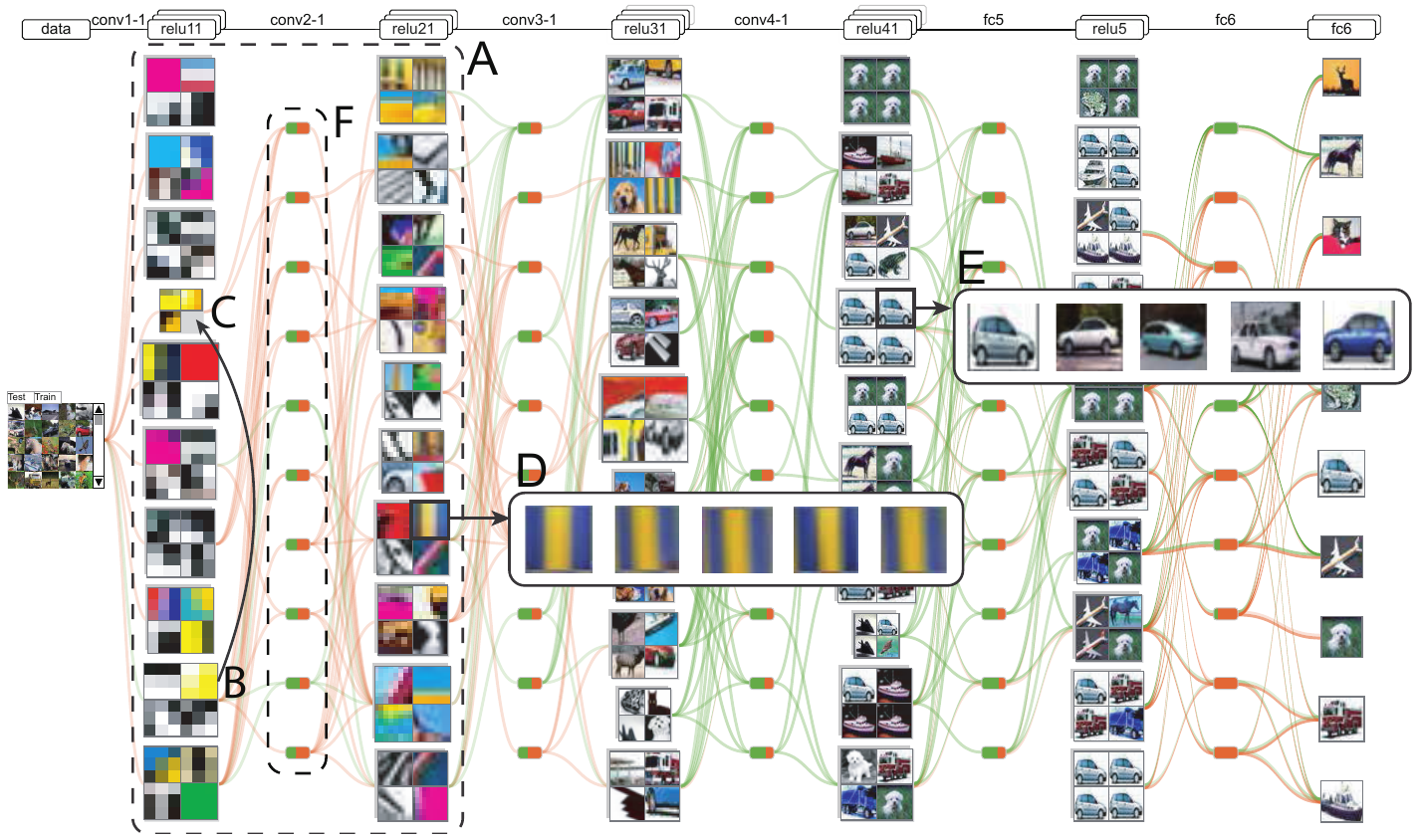}
\caption{CNNVis, a visual analytics approach to understanding and diagnosing deep convolutional neural networks (CNNs)~\cite{Liu2016_TVCG_Towards} with a large number of neurons and connections.}
\label{fig:network-based-large}
\end{figure}

The aforementioned technique can effectively visualize neural networks with several dozens of neurons.
However, as the number of neurons and connections increase, the visualization may become cluttered and difficult to understand~\cite{Tzeng2005_IEEEVisualization_Opening}.
%To solve this problem, Liu et al.~\cite{Liu2016_TVCG_Towards} have developed CNNVis, a visual analytics system that helps machine learning experts understand and diagnose deep convolutional neural networks (CNNs) with thousands of neurons and millions of connections (Fig.~\ref{fig:network-based-large}).
To solve this problem, Liu et \doc{al.~\cite{Liu2016_TVCG_Towards} developed} CNNVis, a visual analytics system that helps machine learning experts understand and diagnose deep convolutional neural networks (CNNs) with thousands of neurons and millions of connections (Fig.~\ref{fig:network-based-large}).
To display large CNNs, the layers and neurons are clustered.
A representative layer (neuron) is selected for each layer (neuron) cluster.
%To effectively display many connections, a biclustering-based algorithm is developed to bundle the edges and reduce visual clutter.
To effectively display many connections, a biclustering-based algorithm is \doc{used} to bundle the edges and reduce visual clutter.
%Moreover, CNNVis supports analysis of interactions between neurons and multiple facets of each neuron.
Moreover, CNNVis supports the analysis of multiple facets of each neuron.
To this end, CNNVis visualizes the learned features of each neuron cluster by using a hierarchical rectangle packing algorithm.
%A matrix reordering algorithm is also developed to reveal the activation patterns of neurons. % based on the Held-Karp algorithm.
A matrix reordering algorithm \doc{was} also developed to reveal the activation patterns of neurons.

\subsection{Diagnosis}
Researchers have developed visual analytics techniques that diagnose model performance for binary classifiers~\cite{Amershi2015_CHI_ModelTracker}, multi-class classifiers~\cite{Liu2016_TVCG_Towards,Alsallakh2014_TVCG_Visual,Ren2016_TVCG_Squares}, and topic models~\cite{Chuang2013_ICML_Topic}.
%The goal of these techniques is to help machine learning experts understand the reason why a training process did not achieve a desirable performance so that they can make better choices (e.g., select better features) for improving the model performance.
The goal of these techniques is to help machine learning experts \doc{understand why} a training process did not achieve a desirable performance so that they can make better choices (e.g., select better features) \doc{to improve} the model performance.
To this end, current techniques utilize the  prediction score distributions of the model (i.e., sample-class probability) to evaluate the error severity and study how the score distributions correlate with misclassification and selected features.
%Here a model's prediction score distributions refer to the probability that a training or test sample belongs to each class predicted by the classifier. %~\cite{Alsallakh2014_TVCG_Visual}.
%Here a model's prediction score distributions refer to how this classifer
% that did not achieve desired performance.

\begin{figure}[t]
\centering
\includegraphics[width=\textwidth]{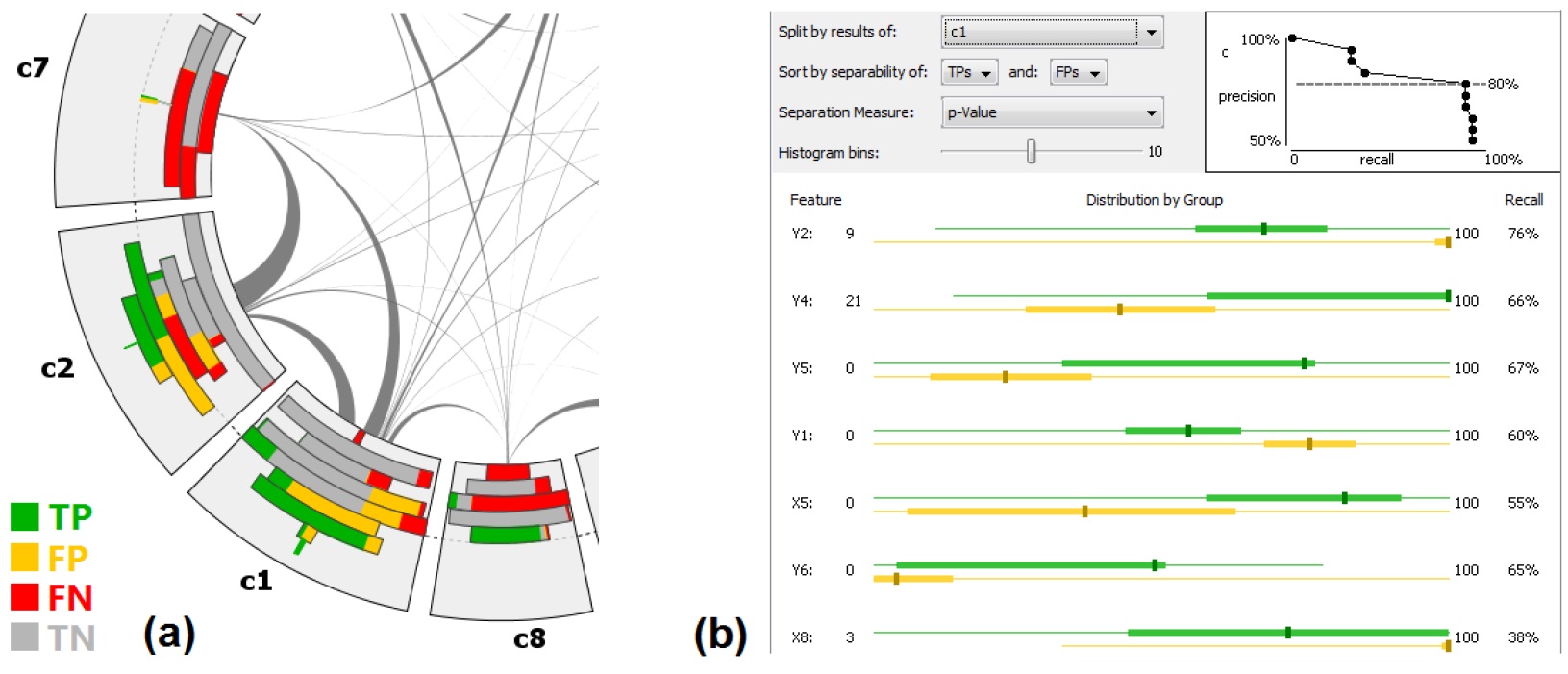}
\caption{
%A visual analytics tool for model performance diagnosis~\cite{Alsallakh2014_TVCG_Visual}.
A visual analytics tool that helps machine learning experts diagnose model performance with (a) a confusion wheel and (b) a feature analysis view~\cite{Alsallakh2014_TVCG_Visual}.
}
\label{fig:confusion-wheel}
\end{figure}

%Fig.~\ref{fig:confusion-wheel} shows a model performance diagnosis tool developed by Alsallakh et al.~\cite{Alsallakh2014_TVCG_Visual}.
One typical example is the model performance diagnosis tool developed by Alsallakh et al.~\cite{Alsallakh2014_TVCG_Visual}.
This tool consists of a confusion wheel (Fig.~\ref{fig:confusion-wheel}(a)) and a feature analysis view (Fig.~\ref{fig:confusion-wheel}(b)).
%For each class $c_i$, the wheel confusion view visualizes the distribution of true-positive (TPs), false-positive (FPs), false-negative (FNs), and true-negative (TNs) samples in the class by using a histogram.
%Each bin in the histogram corresponds to samples with certain prediction scores.
The confusion wheel depicts the prediction score distributions by using histograms.
For each class $c_i$, bins that correspond to samples with low (high) prediction scores of $c_i$ are placed close to the inner (outer) ring.
%The chords in the confusion wheel visualizes the number of samples that belong to class $c_i$ misclassified to class $c_j$ (between-class confusion). %?
The chords in the confusion wheel \doc{visualize} the number of samples that belong to class $c_i$ misclassified to class $c_j$ (between-class confusion).
This view enables users to quickly identify the samples that are misclassified with a low probability (e.g., the false-negative samples (FNs) in $c_7$).
These samples are easier to improve compared with other samples.
The feature analysis view illustrates how two groups of samples (e.g., true-positive samples and false-positive samples) can be separated by using certain features.
This view helps users to make better choices in terms of feature selection.

\begin{figure}[t]
\centering
\includegraphics[width=\textwidth]{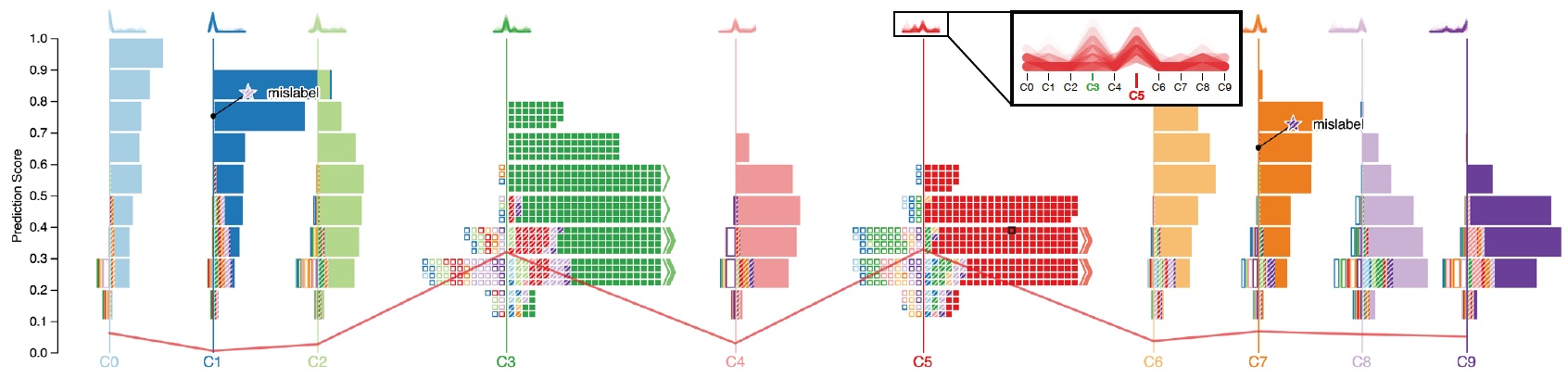}
\caption{
%Squares, a visual analytics tool that supports performance diagnosis of multi-class classifiers within a single visualization to reduce users' cognitive load during analysis~\cite{Ren2016_TVCG_Squares}.
Squares, a visual analytics tool that supports performance diagnosis of multi-class classifiers within a single visualization to reduce \doc{the cognitive load of users} during analysis~\cite{Ren2016_TVCG_Squares}.
}
\label{fig:squares}
\end{figure}

Although the aforementioned technique provides valuable guidance for performance improvement, the confusion wheel can introduce distortion by displaying histograms in a radial display.
Researchers also point out that multiple coordinated visualizations may add complexity to the diagnosis process~\cite{Ren2016_TVCG_Squares}.
%To eliminate the distortion and reduce the users' cognitive load, Ren et al. proposed Squares~\cite{Ren2016_TVCG_Squares}, a visual analytics tool that supports performance diagnosis within a single visualization.
To eliminate the distortion and reduce \doc{the cognitive load of users}, Ren et al. proposed Squares~\cite{Ren2016_TVCG_Squares}, a visual analytics tool that supports performance diagnosis within a single visualization.
%As shown in Fig.~\ref{fig:squares}, Squares is able to show prediction score distributions at multiple levels of details.
As shown in Fig.~\ref{fig:squares}, Squares is able to show prediction score distributions at multiple levels of \doc{detail}.
%The classes expanded to show the lowest level of details (e.g., $c_3$ and $c_5$) are displayed as boxes.
The \doc{classes, when} expanded to show the lowest level of \doc{detail} (e.g., $c_3$ and $c_5$)\doc{,} are displayed as boxes.
%Here, each box represents a (training or test) sample.
\doc{Each} box represents a (training or test) sample.
The color of the box encodes the class label of the corresponding sample and the texture represents whether a sample is classified correctly (solid fill) or not (striped fill).
%The classes with the least details (e.g., $c_0$ and $c_1$) are displayed as stacks.
The classes with the \doc{least number of} details (e.g., $c_0$ and $c_1$) are displayed as stacks.
Squares also allows machine learning experts to explore between-class confusion (see polylines in Fig.~\ref{fig:squares}) within the same visualization.

%More recently, there are some initial efforts on diagnosing deep learning models~\cite{Liu2016_TVCG_Towards, Zahavy2016_ICML_Graying}.
More recently, there \doc{have been} some initial efforts on diagnosing deep learning models~\cite{Liu2016_TVCG_Towards, Zahavy2016_ICML_Graying}.
%For example, Liu et al.~\cite{Liu2016_TVCG_Towards} developed
One example is CNNVis~\cite{Liu2016_TVCG_Towards} (Fig.~\ref{fig:network-based-large}).
By revealing multiple facets of the neurons, the interactions between neurons, and relative weight changes between layers, CNNVis allows machine learning experts to debug a training process that fails to converge or does not achieve an acceptable performance.
%It also helps find potential directions to prevent the training process from getting stuck or improve the model performance.
It also helps \doc{to} find potential directions to prevent the training process from getting stuck or improve the model performance.
%Another example is the method developed by Rauber et al.~\cite{Rauber2016_TVCG_Visualizing}, which enables experts to diagnose whether a class is absent in a layer by disclosing the relationships between neurons and classes.
Another example is the method developed by Zahavy et al.~\cite{Zahavy2016_ICML_Graying}, which employs t-SNE to disclose relationships between learned representations and uses saliency maps to help users analyze influential features.
%Case studies on three ATARI games demonstrate the ability of this method to find problems pertain to game modeling, initial and terminal state modeling, and score over-fitting.
Case studies on three ATARI games demonstrate the ability of this method to find problems \doc{that} pertain to game modeling, initial and terminal state modeling, and score over-fitting.

\subsection{Refinement}

%After understanding how machine learning models behave and why they do not achieve desirable performance, machine learning experts usually wish to refine the model by incorporating the knowledge learned.
After \doc{they gain an understanding of} how machine learning models behave and why they do not achieve \doc{a} desirable performance, machine learning experts usually wish to refine the model by incorporating the knowledge learned.
%To facilitate this process, researchers have developed visual analytics systems that provide interaction capabilities for improving performance of  supervised~\cite{Paiva2015_TVCG_Approach} or unsupervised models~\cite{Wang2016_TVCG_TopicPanorama,Liu2016_TVCG_Uncertainty}.	
To facilitate this process, researchers have developed visual analytics systems that provide interaction capabilities for improving \doc{the} performance of  supervised~\cite{Paiva2015_TVCG_Approach} or unsupervised models~\cite{Wang2016_TVCG_TopicPanorama,Liu2016_TVCG_Uncertainty}.	
%To facilitate this process, researchers have developed visual analytics systems that provide interaction capabilities for improving performance of clustering algorithms~\cite{Tzeng2004_SixthJointEurographicsIEEETCVGConferenceonVisualization_Cluster,Ahmed2011_EuroVAWorkshop_Steerable}, topic models~\cite{Choo2013_TVCG_UTOPIAN}, graph matching~\cite{Wang2016_TVCG_TopicPanorama,Liu2014_VAST_TopicPanorama}, information retrieval~\cite{Liu2016_TVCG_Uncertainty}, and classification~\cite{Paiva2015_TVCG_Approach,Alsallakh2014_TVCG_Visual}.
\begin{figure}[t]
\centering
\includegraphics[width=\textwidth]{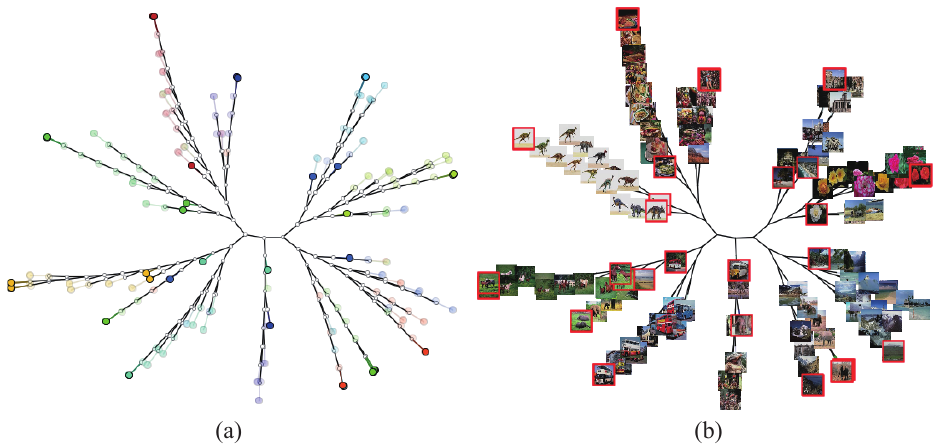}
\caption{Interactive training sample selection that enables classifier refinement~\cite{Paiva2015_TVCG_Approach}. Candidate samples are represented by (a) circles and (b) images.}
\label{fig:refine-supervised}
\end{figure}

Current techniques for refining supervised models mainly focus on multi-class classifiers~\cite{Paiva2015_TVCG_Approach,Alsallakh2014_TVCG_Visual}.
These techniques allow users to insert their knowledge by controlling factors that significantly affect classification results.
Commonly considered factors include training samples, features, types of classifiers, and parameters used in training. %~\cite{Paiva2015_TVCG_Approach}.
%Take the technique developed by Paiva et al.~\cite{Paiva2015_TVCG_Approach} as an example.
For example, the technique developed by Paiva et al.~\cite{Paiva2015_TVCG_Approach} allows users to interactively select training samples, modify their labels, incrementally update the model, and rebuild the model by using new classes.
Fig.~\ref{fig:refine-supervised} shows how this technique supports informed training sample selection.
Here, each sample is displayed as a point in Fig.~\ref{fig:refine-supervised}(a) and an image in Fig.~\ref{fig:refine-supervised}(b).
These samples are organized by using Neighbor Joining trees~\cite{Paiva2011_TVCG_Improved}.  %? reference
%After observing the trees, the user carefully selected 43 samples from the core of the tree and the end of branches.
After observing the trees, the user carefully selected 43 samples from the core of the tree and the end of \doc{the} branches.
Training with these samples generates a classifier with an accuracy of 97.43\%.

%Here, each point in Fig.~\ref{fig:refine-supervised}(a) and each image in Fig.~\ref{fig:refine-supervised}(b) represent a sample that can be used for training.
%An example for training set selection is shown in Fig.~\ref{fig:refine-supervised}.

% shows a visualization generated

\begin{figure}[t]
\centering
\includegraphics[width=\textwidth]{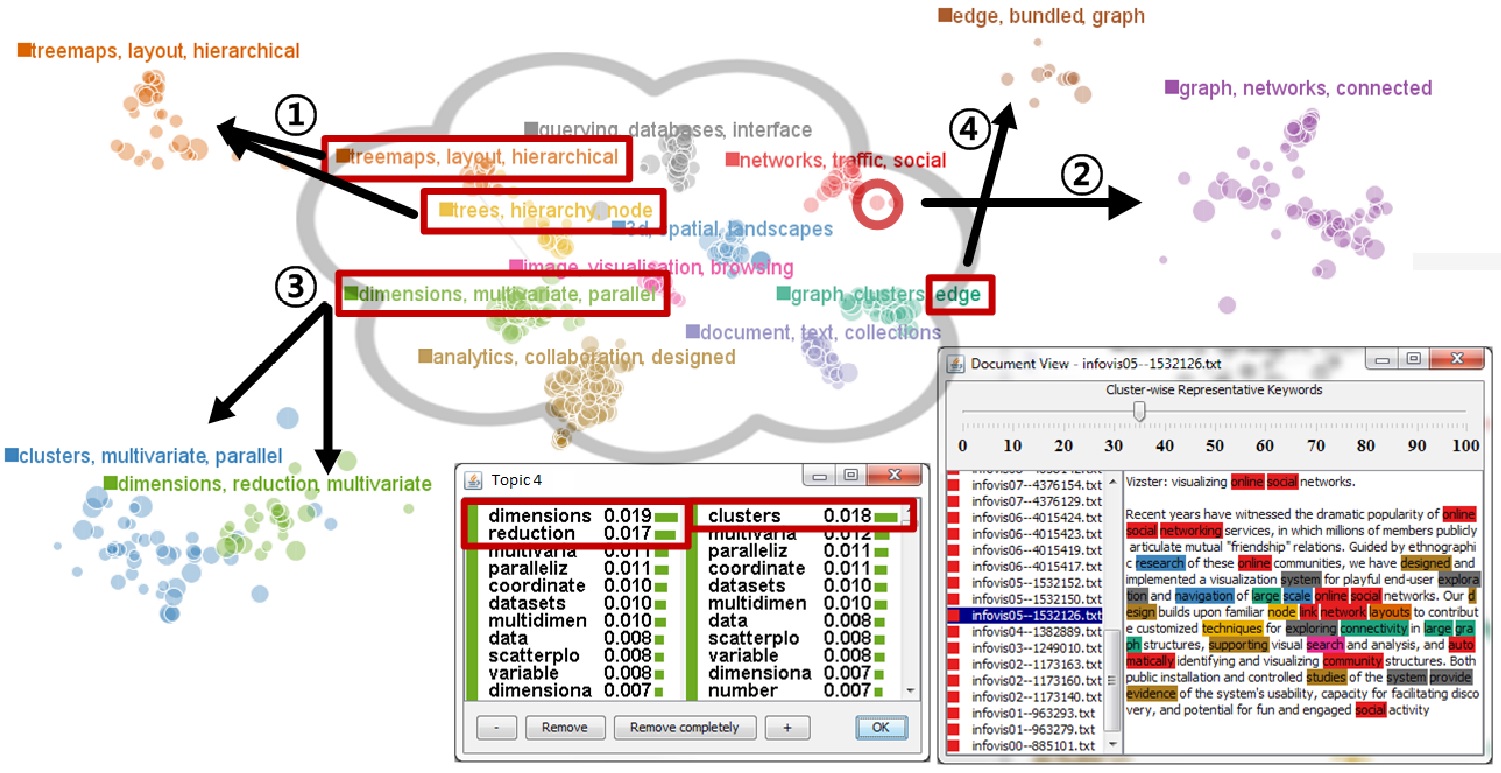}
\caption{UTOPIAN~\cite{Choo2013_TVCG_UTOPIAN}, a visual analytics system for interactive refinement of topic models.}
\label{fig:utopian}
\end{figure}

%Different from refinement techinques for supervised models, many refinement models for unsupervised models
The techniques for refining unsupervised models usually incorporate user knowledge into the model in a semi-supervised manner~\cite{Tzeng2004_SixthJointEurographicsIEEETCVGConferenceonVisualization_Cluster,Choo2013_TVCG_UTOPIAN,Liu2014_VAST_TopicPanorama}.
%A typical work in this field is UTOPIAN~\cite{Choo2013_TVCG_UTOPIAN}, a visual analytics system for refining topic model results.
A typical \doc{example} in this field is UTOPIAN~\cite{Choo2013_TVCG_UTOPIAN}, a visual analytics system for refining topic model results.
%In UTOPIAN, the topics are initially learned by using Nonnegative Matrix Factorization (NMF)~\cite{Lee1999_Nature_Learning}.
%The learned topics are displayed by using a scatterplot visualization.
In UTOPIAN, the topics are initially \doc{learned using} Nonnegative Matrix Factorization (NMF)~\cite{Lee1999_Nature_Learning} \doc{and the} learned topics are \doc{displayed using} a scatterplot visualization.
As shown in Fig.~\ref{fig:utopian}, UTOPIAN allows users to interactively (1) merge topics, (2) create topics based on exemplar documents, (3) split topics, and (4) create topics based on keywords.
Moreover, UTOPIAN also supports topic keyword refinement.
%All these interactions are centered around a semi-supervised formulation of NMF that enables an easy incorporation of user knowledge and incremental update of the topic model.  %?
All these interactions are centered around a semi-supervised formulation of NMF that enables an easy incorporation of user knowledge and \doc{an} incremental update of the topic model.

%TopicPanorama
%There are also some refinement tools that aim to help business professionals who are not familiar with those complex machine learning models.
There are also some refinement tools that aim to help business professionals who are not familiar \doc{with complex} machine learning models.
For example, Wang et al. developed a visual analytics system, TopicPanorama~\cite{Wang2016_TVCG_TopicPanorama,Liu2014_VAST_TopicPanorama}, to help business professionals analyze and refine a full picture of relevant topics discussed in multiple textual sources. %(Fig.~\ref{fig:topicpanorama}(a))
%Fig.~\ref{fig:topicpanorama}(a) shows a full picture generated by TopicPanorama.
The full picture is generated by matching the topic graphs extracted from different sources with a scalable algorithm to learn correlated topic models~\cite{Chen-CTM:nips13}.
TopicPanorama allows users to identify potentially incorrect matches by examining the uncertainties of the matches.
Moreover, by incorporating metric learning and feature selection into the graph matching model, TopicPanorama allows users to incrementally improve and refine the matching model.

\begin{figure}[t]
\centering
\includegraphics[width=\textwidth]{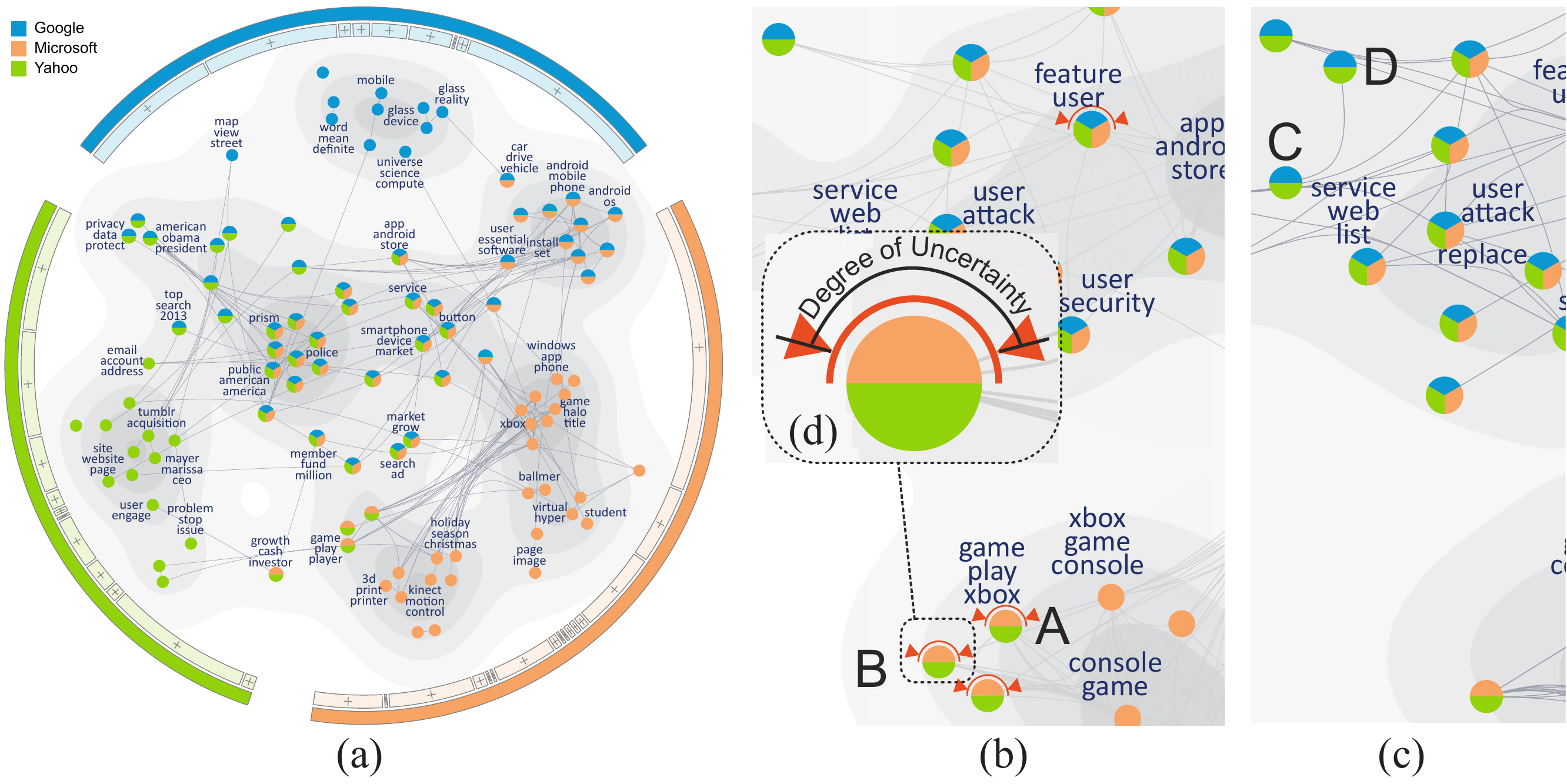}
\caption{TopicPanorama~\cite{Wang2016_TVCG_TopicPanorama}, a visual analytics system for analyzing a full picture of relevant topics from multiple sources: (a) Panorama visualization, (b) a matching result with two incorrect matches \textbf{A} and \textbf{B}, (c) the updated matching result with corrected matches \textbf{C} and \textbf{D}, and (d) an uncertainty glyph.}
\label{fig:topicpanorama}
\end{figure}

Fig.~\ref{fig:topicpanorama}(a) shows a full picture of the topics related to three IT companies: Google, Microsoft, and Yahoo.
%Here, the topic nodes of different companies (sources) are represented by different colors and the common topics are encoded by a pie chart.
Here, the topic nodes of different companies (sources) are represented \doc{with} different colors and the common topics are encoded \doc{in} a pie chart.
%A public relation manager cared about game related topics, thus she checked potential incorrect matches .
%After examining the uncertainty glyphs (Fig.~\ref{fig:topicpanorama}(d)), she found two incorrect matchings \textbf{A} and \textbf{B} (Fig.~\ref{fig:topicpanorama}(b)).
%A public relation manager cared about game related topics, thus she enabled the uncertainty glyphs (Fig.~\ref{fig:topicpanorama}(d)) to examine potential incorrect matches.
A public \doc{relations} manager cared about game related topics, \doc{so} she enabled the uncertainty glyphs (Fig.~\ref{fig:topicpanorama}(d)) to examine potential incorrect matches.
%After some exploration, she identified two incorrect matches \textbf{A} and \textbf{B} that match Microsoft Xbox games to Yahoo sport games (Fig.~\ref{fig:topicpanorama}(b)).
After some exploration, she identified two incorrect \doc{matches,} \textbf{A} and \doc{\textbf{B},} that match Microsoft Xbox games to Yahoo sport games (Fig.~\ref{fig:topicpanorama}(b)).
%After she unmatched \textbf{B}, she found \textbf{A} was changed to \textbf{C} and \textbf{B} was changed to \textbf{D}, which correctly match Google sport games to Yahoo sport games (Fig.~\ref{fig:topicpanorama}(c)).
After she unmatched \textbf{B}, she found \textbf{A} was changed to \textbf{C} and \textbf{B} was changed to \textbf{D}, which correctly \doc{matched} Google sport games to Yahoo sport games (Fig.~\ref{fig:topicpanorama}(c)).

%topic nodes of different sources are encoded by
%A major feature of

%Here, each color encodes a source
%such as public relation managers
\begin{figure}[t]
\centering
\includegraphics[width=\textwidth]{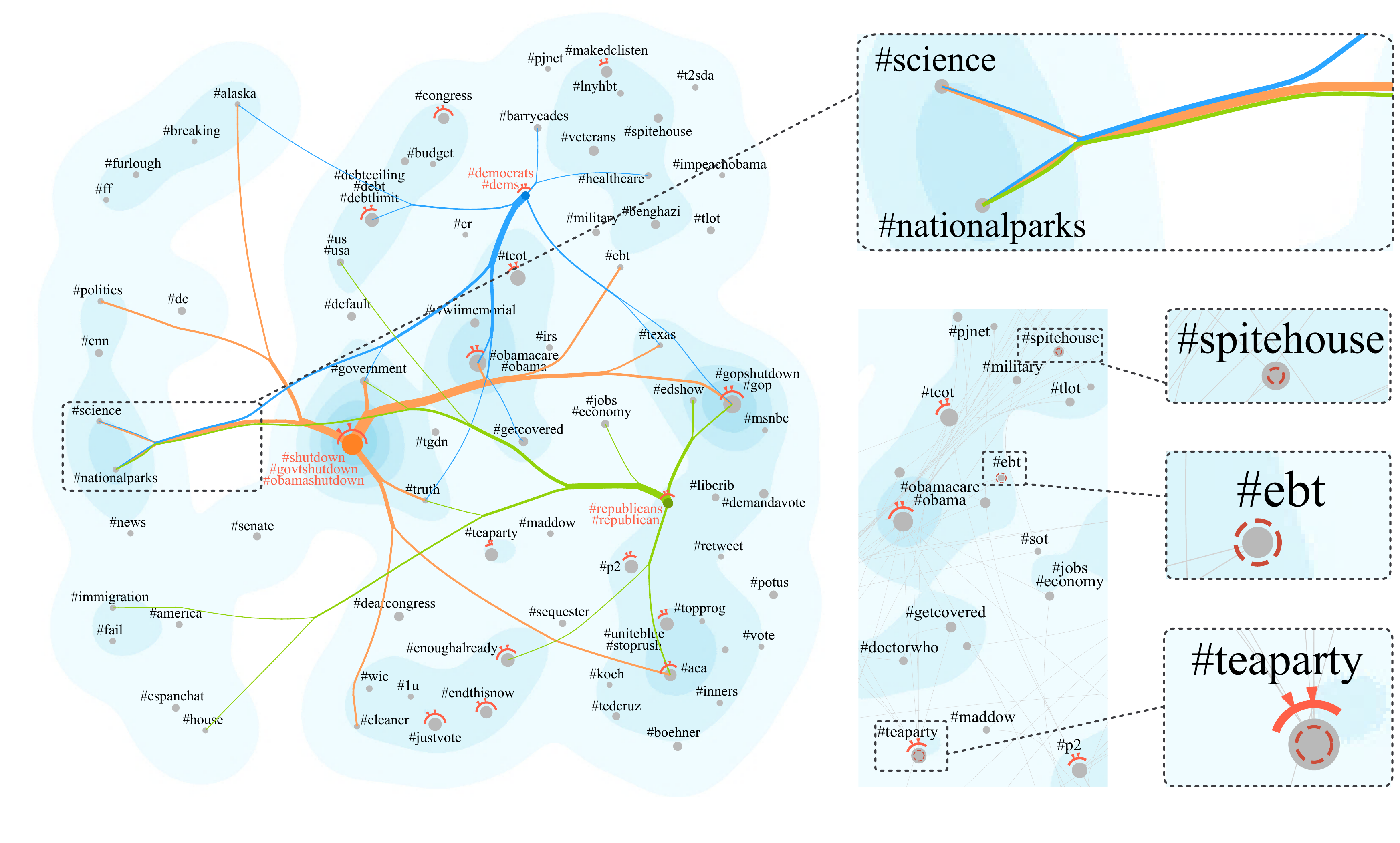}
\caption{
%MutualRanker~\cite{Liu2016_TVCG_Uncertainty}, a visual analytics toolkit to retrieve salient posts, users and hashtags.
MutualRanker~\cite{Liu2016_TVCG_Uncertainty}, a visual analytics toolkit to retrieve salient posts, \doc{users,} and hashtags.
%MutualRanker enables interactive refinement of the uncertain results.
MutualRanker enables interactive refinement \doc{of uncertain} results.
}
\label{fig:mutualranker}
\end{figure}

%For example, Liu et al.~\cite{Liu2016_TVCG_Uncertainty} developed MutualRanker, a visual analytics toolkit to retrieve salient posts, users and hashtags.
%Another example is MutualRanker~\cite{Liu2016_TVCG_Uncertainty}, a visual analytics toolkit to retrieve salient posts, users and hashtags.
%To effectively retrieve salient posts, users and hashtags, they build a mutual reinforcement graph model~\cite{Wei2008_SIGIR_Query} that jointly considers the content quality of posts, social influence of users, and the popularity of hashtags.
%
%%To help business professionals understand and interactively refine the retrieved data
%on that visually illustrates the posts, users, hashtags, their relationships, and the uncertainty in the results.
%By this visualization, business professionals are able to detect the most uncertain results and interactively refine the results.
%A typical use case is shown in Fig.~\ref{fig:mutualranker}, where an expert finds the cluster ``nationalparks'' shares the uncertainty propagated from the ``shutdown'', ``democrats'', and ``republicans'' cluster.
%It indicates there is high uncertainty in the ranking scores of the hashtags in the ``nationalparks'' cluster.
%He then refines the results according to his domain knowledge.

%Another example is MutualRanker~\cite{Liu2016_TVCG_Uncertainty}, a visual analytics tool to retrieve salient posts, users and hashtags.
Another example is MutualRanker~\cite{Liu2016_TVCG_Uncertainty}, a visual analytics tool to retrieve salient posts, \doc{users,} and hashtags.
%To effectively retrieve salient posts, users and hashtags, they build a mutual reinforcement graph (MRG) model~\cite{Wei2008_SIGIR_Query} that jointly considers the content quality of posts, social influence of users, and the popularity of hashtags.
To effectively retrieve salient posts, users and hashtags, they \doc{built} a mutual reinforcement graph (MRG) model~\cite{Wei2008_SIGIR_Query} that jointly considers the content quality of posts, \doc{the} social influence of users, and the popularity of hashtags.
They also analyzed the uncertainty in the results.
Based on the retrieved data and the uncertainty, they developed a composite visualization that visually illustrates the posts, users, hashtags, their relationships, and the uncertainty in the results.
%By this visualization, business professionals are able to easily detect the most uncertain results and interactively refine the MRG model.
\doc{With} this visualization, business professionals are able to easily detect the most uncertain results and interactively refine the MRG model.
To efficiently refine the model, they developed a random-walk-based Monte Carlo sampling method that can locally update the model based on user interactions.
%A typical use case of MutualRanker is shown in Fig.~\ref{fig:mutualranker}, where an expert found that the cluster ``nationalparks'' shared the uncertainty propagated from the ``shutdown'', ``democrats'', and ``republicans'' cluster.
A typical use case of MutualRanker is shown in Fig.~\ref{fig:mutualranker}, where an expert found that the cluster ``nationalparks'' shared the uncertainty propagated from the \doc{``shutdown,'' ``democrats,''} and ``republicans'' cluster.
%It indicates there is high uncertainty in the ranking scores of the hashtags in the ``nationalparks'' cluster.
\doc{This} indicates there is high uncertainty in the ranking scores of the hashtags in the ``nationalparks'' cluster.
%<<<<<<< .mine
%%He then refines the model according to his domain knowledge.
%As the closing of the national parks was a result of the government shutdown and stimulated discussion on Twitter, he
%increased the ranking score of \#nationalparks.
%After the adjustment, the expert noticed the scores of another two hashtag clusters were automatically increased: ¡°\#spitehouse¡± and ¡°\#teaparty.¡±
%=======
According to his domain knowledge, the expert increased the ranking scores of ``\#nationalparks'' in that cluster and the ranking scores of other relevant hashtags were automatically updated.

%A key feature of MutualRanker is that it enables business professionals to detect the most uncertain retrieval results and interactively refine the results.
%It is achieved by modeling and visualizing the uncertainty and its propagation.

% based on semi-supervised Nonnegative Matrix Factorization (NMF)~\cite{Lee1999_Nature_Learning}.

%These methods can be divided into two groups based on the model they refine is unsupervised~\cite{Wang2016_TVCG_TopicPanorama,Liu2016_TVCG_Uncertainty} or supervised~\cite{Paiva2015_TVCG_Approach}.

%% file: challenges.tex
\section{Research Opportunities}

%We regard the existing work as an initial step and there are many research opportunities to be further explored and pursued, which will be discussed in the following subsection in terms of technical challenges and future work.
We \doc{regard existing methods} as an initial step and there are many research opportunities to be further explored and pursued, which will be discussed in the following subsection in terms of technical challenges and future \doc{research}.

\subsection{Creating Explainable Models}
%Although machine learning models are widely used in many applications, they often failed to explain their decisions and actions to users.
Although machine learning models are widely used in many applications, they often \doc{fail} to explain their decisions and actions to users.
%Without a clear understanding, it may hard for the users incorporate their knowledge into the learning process and achieve a better learning performance (e.g., prediction accuracy).
Without a clear understanding, it may \doc{be} hard \doc{for users to} incorporate their knowledge into the learning process and achieve a better learning performance (e.g., prediction accuracy).
As a result, it is desirable to develop more explainable machine learning models, which have the ability to explain their rationale and convey an understanding of how they behave in the learning process.
%The key challenge here is to design an explanation mechanism that is tightly integrated with a machine learning model.
The key challenge here is to design an explanation mechanism that is tightly integrated \doc{into the} machine learning model.

%Accordingly, one interesting future work is to discover which part(s) in the model structure explains its different functions and play a major role to the performance improvement or decline in each iteration.
Accordingly, one interesting future work is to discover which part(s) in the model structure explains its different functions and play a major role \doc{in} the performance improvement or decline \doc{of} each iteration.
%Another interesting venue for future work is to better illustrate rationale behind the model and decision made.
Another interesting venue for future work is to better illustrate \doc{the} rationale behind the model and \doc{the decisions} made.
%Recently, there are some initial efforts in this direction~\cite{Letham2015_Interpretable, Lake2015_Science_Human}.
Recently, there \doc{have been} some initial efforts in this direction~\cite{Letham2015_Interpretable, Lake2015_Science_Human}.
For example, Lake et al.~\cite{Lake2015_Science_Human} developed a probabilistic program induction algorithm.
They built simple stochastic programs to represent concepts, building them compositionally from parts, subparts, and spatial relations.
%They also demonstrated that their algorithm achieved human-level performance on one-shot classification task, while outperforming recent deep learning approaches.
They also demonstrated that their algorithm achieved human-level performance on \doc{a} one-shot classification task, while outperforming recent deep learning approaches.
%However, for the tasks that have abundant training data, such as object and speech recognition, the less explainable deep learning still outperforms their algorithm.
However, for the tasks that have abundant training data, such as object and speech recognition, the less explainable deep learning still outperforms \doc{the} algorithm.
%Thus, there is still a long way ahead for researchers to develop more explainable models on these tasks.
Thus, there is still a long way \doc{to go} for researchers to develop more explainable models \doc{for} these tasks.

%Produce more explainable models, while maintaining a high level of learning performance (prediction accuracy); and
%Enable human users to understand, appropriately trust, and effectively manage the emerging generation of artificially intelligent partners.

%\subsection{Analysis of Unsupervised Models}
%Most of the existing work focuses on analyzing supervised machine learning models.
%
%
%underfitting, overfitting
%
%Ideally, you want to select a model at the sweet spot between underfitting and overfitting.
%This is the goal, but is very difficult to do in practice.
%%http://machinelearningmastery.com/overfitting-and-underfitting-with-machine-learning-algorithms/
%
%
%understanding the learned representation of each component
%
%we can give a
%qualitative understanding of learned policies that can help
%the outer optimization loop and help in the debugging process.

\subsection{Analysis of Online Training Process}

%Most of the existing methods focuses on analyzing the final results~\cite{Ren2016_TVCG_Squares} or one snapshot~\cite{Liu2016_TVCG_Towards} of the model in the interactive training process.
Most of the existing methods \doc{focus} on analyzing the final results~\cite{Ren2016_TVCG_Squares} or one snapshot~\cite{Liu2016_TVCG_Towards} of the model in the interactive training process.
%In many cases, only analyzing the results or one snapshot is not enough to understand why a training process did not achieve a desirable performance.
In many cases, only analyzing the results or \doc{a single} snapshot is not enough to understand why a training process did not achieve a desirable performance.
%Thus, it is desirable to analyze the online training process.
Thus, it is \doc{necessary} to analyze the online training process.

One challenge in analyzing the online training process is the difficulty of selecting and comparing representative snapshots from a large number of snapshots.
When comparing different snapshots, one possible solution is to adopt progressive visual analytics~\cite{Stolper2014_TVCG_Progressive} to shorten the period of time between user interactions and the execution of the model.
%The basic idea of progressive visual analytics is producing meaningful partial results during the training process and integrating these partial results into interactive visualization, which allows users to immediately explore these partial results.
The basic idea of progressive visual analytics is \doc{to produce} meaningful partial results during the training process and integrating these partial results into \doc{an} interactive visualization, which allows users to immediately explore \doc{the} partial results.

Another challenge is automatically and accurately detecting anomalies in the training process.
%Nowadays, the training process is sometimes too long (e.g., more than one week for a large deep neural network~\cite{Krizhevsky2012_NIPS_Imagenet}) for an expert to supervise the whole training process.
\doc{Currently}, the training process is sometimes too long (e.g., more than one week for \doc{an expert to supervise the whole training process of} a large deep neural network~\cite{Krizhevsky2012_NIPS_Imagenet}).
%In these scenarios, automatically detecting anomalies and timely notify the expert are necessary.
In these scenarios, \doc{it is necessary to} automatically \doc{detect} anomalies and timely notify \doc{the expert}.
% because it enables experts to switch to other tasks and notify them when detecting an anomaly.
Automatic and accurate identification of anomalies is still a challenging research topic~\cite{Tam2016_TVCG_Analysis}.
%Thus, it is desirable to employ interaction visualization, which can better combine the humans' ability of detecting anomalies and the power of machines to process large amounts of data, which has been initially studied by some recent work~\cite{Cao2015_TVCG_TargetVue, Zhao2014_TVCG_FluxFlow}.\looseness=-1
Thus, it is desirable to employ \doc{an} interactive visualization, which can better combine the \doc{human} ability \doc{to detect} anomalies and the power of machines to process large amounts of data, which has been initially studied \doc{in} some recent work~\cite{Cao2015_TVCG_TargetVue, Zhao2014_TVCG_FluxFlow}.

\subsection{Mixed Initiative Guidance}

%To improve the performance of machine learning models and better incorporate experts' knowledge, researchers have developed a set of guidance techniques.
To improve the performance of machine learning models and better incorporate \doc{the knowledge of experts}, researchers have developed a set of guidance techniques.
%Such efforts are mainly from two research communities: machine learning and information visualization.
Such efforts \doc{have arisen} from two \doc{main} research communities: machine learning and information visualization.
%From the machine learning community, researchers have developed a wide array of techniques on system initiated guidance~\cite{Settles2012_Book_Active,Cohn1994_ML_Improving,Cohn1996_JAIR_Active,Mccallumzy1998_ICML_Employing},
From the machine learning community, researchers have developed a wide array of techniques \doc{for} system initiated guidance~\cite{Settles2012_Book_Active,Cohn1994_ML_Improving,Cohn1996_JAIR_Active,Mccallumzy1998_ICML_Employing},
%where the system would play a more active role, for example, making suggestions about appropriate views or next steps in the iterative and progressive analysis
where the system \doc{plays} a more active role, for example, \doc{by} making suggestions about appropriate views or next steps in the iterative and progressive analysis
process.
%From the information visualization community, researchers have designed a number of techniques on user initiative guidance~\cite{Wang2016_TVCG_TopicPanorama,Liu2016_TVCG_Uncertainty,Choo2013_TVCG_UTOPIAN,Liu2014_VAST_TopicPanorama,Pezzotti2016_TVCG_Approximated},
From the information visualization community, researchers have designed a number of techniques \doc{for} user initiative guidance~\cite{Wang2016_TVCG_TopicPanorama,Liu2016_TVCG_Uncertainty,Choo2013_TVCG_UTOPIAN,Liu2014_VAST_TopicPanorama,Pezzotti2016_TVCG_Approximated},
where the user is the active participant in improving and refining the performance and learning results.

%In many tasks, it is preferable to combine system imitative guidance and user initiative guidance as mixed initiative guidance to maximize the values of both.
In many tasks, it is preferable to combine system imitative guidance and user initiative guidance as mixed initiative guidance to maximize the \doc{value} of both.
Accordingly, mixed initiative guidance is defined as a type of visual reasoning or feedback process in which the human analyst and the machine learning system can both actively foster the guidance to improve the machine learning model.
%Although mixed initiative guidance is very useful, support it is technically demanding.
Although mixed initiative guidance is very useful, \doc{supporting} it is technically demanding.
There are two major challenges that we need to address.

First, it is not easy to seamlessly integrate system initiative guidance and user initiative guidance in one unified framework.
System initiative guidance is usually based on the learning process and the evaluation of the results,
while user initiative guidance is typically based on the experience and domain knowledge of the expert.
%Accordingly, we need to study how to define an efficient working mechanism to integrate them and support smooth communications between them.
Accordingly, we need to study how to define an efficient working mechanism to integrate them and support smooth \doc{communication} between them.
For example, one interesting research problem is how to reveal the provenance of system initiative guidance to illustrate why a suggestion is made by the system.
%Then based on this, the expert can better understand the rationale behind the suggestion and provide his/her feedback accordingly.
\doc{Then,} based on this, the expert can better understand the rationale behind the suggestion and provide his/her feedback accordingly.
%Another potential research problem is to effectively extract appropriate and sufficient user/system data to create a unified model for both the user and system.
Another potential research problem is to effectively extract appropriate and sufficient user/system data to create a unified model for both the user and \doc{the} system.

%Second, there may exist several conflicts between system initiative guidance and user initiative guidance in real-world applications.
Second, there may \doc{be} several conflicts between system initiative guidance and user initiative guidance in real-world applications.
%For example, for a given training sample, the system and the user may have different opinion on which class it belongs to.
For example, for a given training sample, the system and the user may have different \doc{opinions} on which class it belongs to.
As a result, how to solve these conflicts is an interesting research problem \doc{that needs further exploration}.
%traditional: system-initiative
%new: combine  user-initiative, system-initiative

\subsection{Uncertainty}

%While visual analytics is very useful in helping machine learning experts gain insights into the working mechanisms of the models and improve model performance, it may also introduce uncertainties into the analysis process.
While visual analytics is very useful in helping machine learning experts gain insights into the working mechanisms \doc{of models} and \doc{devise ways to} improve model performance, it may also introduce uncertainties into the analysis process.
%It has been shown that the uncertainty awareness positively influences human trust building and decision making~\cite{Sacha2016_TVCG_Role}.
It has been shown \doc{that uncertainty} awareness positively influences human trust building and decision making~\cite{Sacha2016_TVCG_Role}.
%Thus, it is important to quantify and analyze uncertainties~\cite{Correa2009_VAST_framework} in interactive model analysis, which is challenging due to the following reasons.
Thus, it is important to quantify and analyze uncertainties~\cite{Correa2009_VAST_framework, Wu2010_TVCG_opinionseer} in interactive model analysis, which is challenging \doc{for a number of} reasons.
%However, handling uncertainties in interactive model analysis is difficult.
%Uncertainties may originate from both the machine side (e.g., machine learning models) and the human side (e.g., incorrect expert feedback) and there may be interplay between uncertainties from the two sides.
%The challenges are detailed as follows.
%The first challenge is to quantify the uncertainties that comes from the machine side.

%First, it is difficult to effectively quantify and display the uncertainties introduced during each step of interactive model analysis.
%First, it is difficult to effectively quantify and display the uncertainties introduced during interactive model analysis.
%Uncertainties, originating from each step of the analysis (e.g., learning, visualization, and refinement), will propogate and aggregate during the whole process.
First, uncertainties may originate from each stage of the interactive model analysis process (e.g., training, visualization, refinement) and increase, decrease, split, and merge during the whole process~\cite{Wu2012_TVCG_Visualizing}.
%For example, the uncertainty of missing data may be enlarged as they
%Thus, it is difficult to effectively quantify the uncertainties of each single stage.
Thus, it is difficult to effectively quantify the uncertainties.
%One interesting direction for future work is to develop visual analytics techniques to effectively measure and quantify the uncertainty in data processing, model building, and visualization~\cite{Sacha2016_TVCG_Role} and help experts quickly identify the potential issues in a machine learning model of interest.
One interesting direction for future \doc{research} is to develop visual analytics techniques \doc{that} effectively measure and quantify the uncertainty in data processing, model building, and visualization~\cite{Sacha2016_TVCG_Role} and help experts quickly identify the potential issues in a machine learning model of interest.\looseness=-1
%One interesting problem is to leverage existing attempts to measure uncertainties~\cite{Sacha2016_TVCG_Role} and display uncertainty propagation~\cite{Wu2012_TVCG_Visualizing} to help machine learning experts make better decisions.
% of data processing, modeling, and visualization stage

%In interactive model analysis, uncertainties may originate from both the machine side (i.e., data, machine learning models, visualization) and the human side (e.g., incorrect expert feedback).

%Second, it is challenging to unifiedly model uncertainties from the machine side, uncertainties from the human side, and their interactions.
Second, it is challenging to model different types of uncertainties as well as their interactions by using a unified framework.
%Second, it is challenging to model interactions between different types of uncertainties.
%in interactive model analysis
During the interactive model analysis process, there are uncertainties that originate from the machine side (e.g., imperfect machine learning models) and uncertainties that originate from the human side (e.g., incorrect expert feedback).
%These two kinds of uncertainties will interact and influence with each other.
These two kinds of uncertainties will interact \doc{with} and \doc{influence each} other.
%For example, if the system presents misleading information to the experts, they may return incorrect feedbacks that result in problematic modification of the model.
For example, if the system presents misleading information to the experts, they may return incorrect \doc{feedback} that \doc{results} in problematic modification of the model.
Another example is that allowing experts to view and refine results of many test samples may encourage overfitting~\cite{Ren2016_TVCG_Squares}.
%Thus, it is important to take into account interactions between different types of uncertainties and.
Accordingly, an interesting research problem is how to model different types of uncertainties and their interactions with a unified model.\looseness=-1
% from the machine side, uncertainties from the human side, and their interactions.
%there may be interplay between uncertainties from the machine side and uncertainties from the human side.

%More specifically, we are faced with the following three challenges.
%Handling uncertainties is very important since
%Uncertainties may come from the machine side as well as the human side.